\documentclass[sigconf]{acmart}
\usepackage{multirow}
\usepackage{colortbl}
\usepackage{makecell}
\usepackage{natbib}
\usepackage{enumitem}

\AtBeginDocument{%
  }

\setcopyright{acmlicensed}
\copyrightyear{2025}
\acmYear{2025}
\acmDOI{XXXXXXX.XXXXXXX}
\acmConference[MedicalAI 2025]{The Conference on MedicalAI}{Jan. 1--Dec. 31, 2025}{Seoul, Korea}
\acmISBN{978-1-4503-XXXX-X/18/06}

\begin{document}

\title{CoFE: A Framework Generating Counterfactual ECG for Explainable Cardiac AI-Diagnostics}

\author{Jong-Hwan Jang, Junho Song, and Yong-Yeon Jo}
\email{{jangood1122, jhsong, yy.jo}@medicalai.com}
\affiliation{%
  \institution{MedicalAI Co., Ltd.}
  \city{Seoul}
  \country{Korea}
}
% \author{Jong-Hwan Jang}
% \email{jangood1122@medicalai.com}
% \affiliation{
%   \institution{AI Group, Medical AI Co., Ltd.}
%   \city{Seoul}
%   \country{Republic of Korea}
% }
% \author{Junho Song}
% \email{jhsong@medicalai.com}
% \affiliation{
%   \institution{AI Group, Medical AI Co., Ltd.}
%   \city{Seoul}
%   \country{Republic of Korea}
% }
% \author{Joon-myoung Kwon}
% \email{cto@medicalai.com}
% \affiliation{
%   \institution{AI Group, Medical AI Co., Ltd.}
%   \city{Seoul}
%   \country{Republic of Korea}
% }
% \author{Yong-Yeon Jo}
% \email{yy.jo@medicalai.com}
% \affiliation{
%   \institution{AI Group, Medical AI Co., Ltd.}
%   \city{Seoul}
%   \country{Republic of Korea}
% }

\renewcommand{\shortauthors}{Jong-Hwan Jang et al.}

\begin{abstract}
Recognizing the need for explainable AI (XAI) approaches to enable the successful integration of AI-based ECG prediction models (AI-ECG) into clinical practice, we introduce a framework generating \textbf{Co}unter\textbf{F}actual \textbf{E}CGs (i,e., named CoFE) to illustrate how specific features, such as amplitudes and intervals, influence the model's predictive decisions. To demonstrate the applicability of the CoFE, we present two case studies: atrial fibrillation classification and potassium level regression models. The CoFE reveals feature changes in ECG signals that align with the established clinical knowledge. 
By clarifying both \textbf{where valid features appear} in the ECG and \textbf{how they influence the model’s predictions}, we anticipate that our framework will enhance the interpretability of AI-ECG models and support more effective clinical decision-making.
Our demonstration video is available at: \url{https://www.youtube.com/watch?v=YoW0bNBPglQ}.
% By clarifying both \textbf{where} and \textbf{how} certain features matter, we anticipate that our framework will enhance the interpretability of AI-ECG models and support more effective clinical decision-making.
\end{abstract}

\begin{CCSXML}
<ccs2012>
   <concept>
       <concept_id>10010405.10010444.10010449</concept_id>
       <concept_desc>Applied computing~Health informatics</concept_desc>
       <concept_significance>500</concept_significance>
       </concept>
 </ccs2012>
\end{CCSXML}

\ccsdesc[500]{Applied computing~Health informatics}

\keywords{Explainable AI, Generative AI, Counterfactual explanation, Electrocardiogram, Biosignal}

\maketitle

\section{INTRODUCTION}

\begin{figure}[!h]
    \centering
    \includegraphics[width=\linewidth]{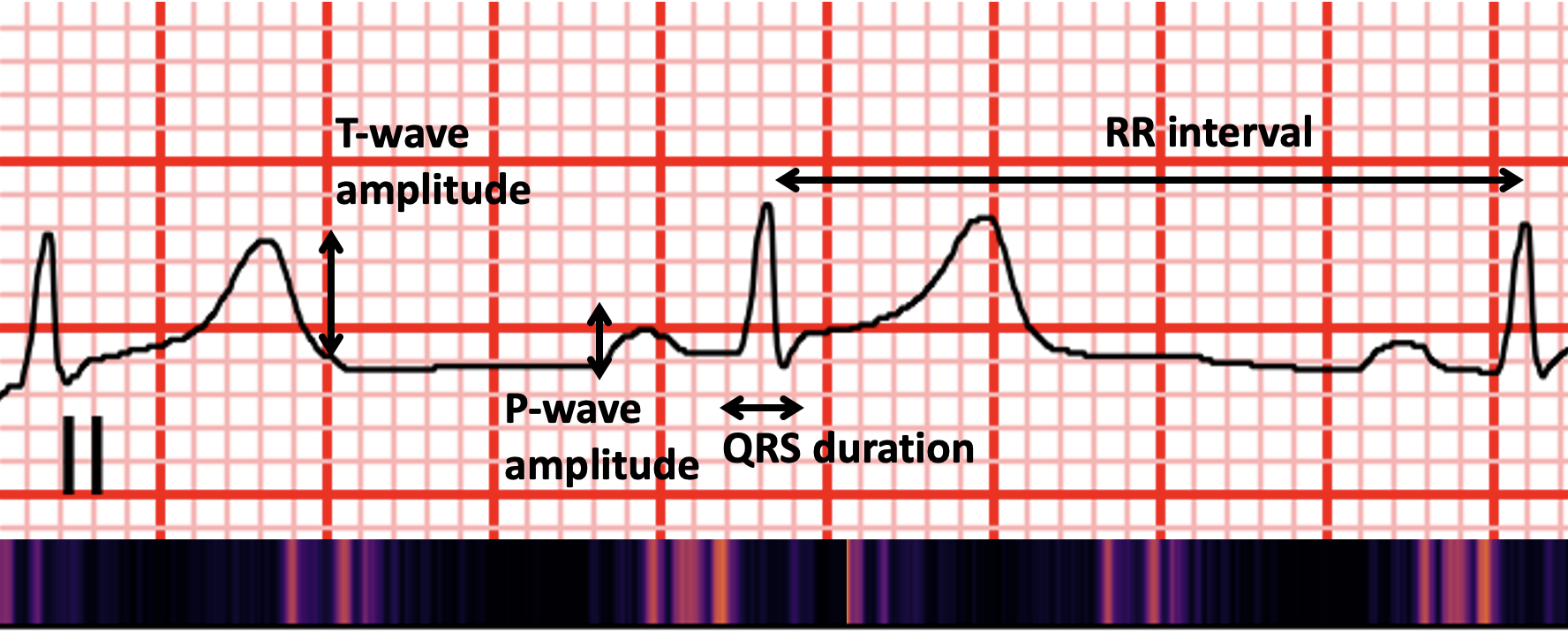}
    \caption{A representative saliency map on an ECG highlights the regions the AI-ECG attends to during prediction. While such maps provide visual insight into its focus, they offer limited explanation of the decision-making process or how predictions change with perturbations in features.}
    \label{fig:ECGreport}
\end{figure}

%% XAI 중요성 및 현재 ECG XAI의 한계
Explainable Artificial Intelligence (XAI) becomes increasingly important in clarifying how AI-based ECG prediction models (AI-ECG) reach their predictions, as it enhances reliability for both clinicians and patients~\cite{attia2021application}. 
Typically, attribution-based approaches such as Saliency Map~\cite{niebur2007saliency} and GradCAM\cite{selvaraju2017grad} are widely used as XAI methods~\cite{jo2021detection}. 
As shown in Figure~\ref{fig:ECGreport}, these methods show \textbf{where} the AI-ECG is focusing; however, they are insufficient to explain \textbf{how} the predictions of AI-ECGs are made and change in response to changes in features.
% of ECGs as illustrated in~\ref{fig:ECGreport}.
% As illustrated in~\ref{fig:ECGreport}, the features of ECG are represented with two axes: time interval and amplitude. XAI for ECG is required to show how specific changes in these features influence the AI-ECG's outputs. However, attribution-based methods alone limit to explain the underlying changes for these highlighted features. 

% Introduction of Counterfactual XAI and its benefits
To overcome this limitation, we introduce a Framework generating \textbf{Co}unter\textbf{F}actual \textbf{E}CG for explainable cardiac AI-diagnostics, named CoFE, which complements existing attribution methods (i.e., saliency map) by exploring "what-if" (i.e., counterfactual ECG) scenarios~\cite{verma2020counterfactual}. CoFE represents how the parts of the input ECG that AI-ECG is focusing on change based on the probabilities of AI-ECG. Thus, CoFE provides clinicians with the ability to understand not only where the AI-ECG focuses but also how adjustments to specific ECG intervals or amplitudes could change a prediction. 

This integrated view provides a more detailed explanation, showing clinicians both \textbf{where} and \textbf{how} specific ECG features influence the AI-ECG's predictions.
By presenting the original and counterfactual ECGs side by side, clinicians can visualize potential causes for an AI-ECG's decision, making its reasoning more transparent and clinically meaningful.

\begin{figure*}[!h]
    \centering
    \includegraphics[width=\textwidth]{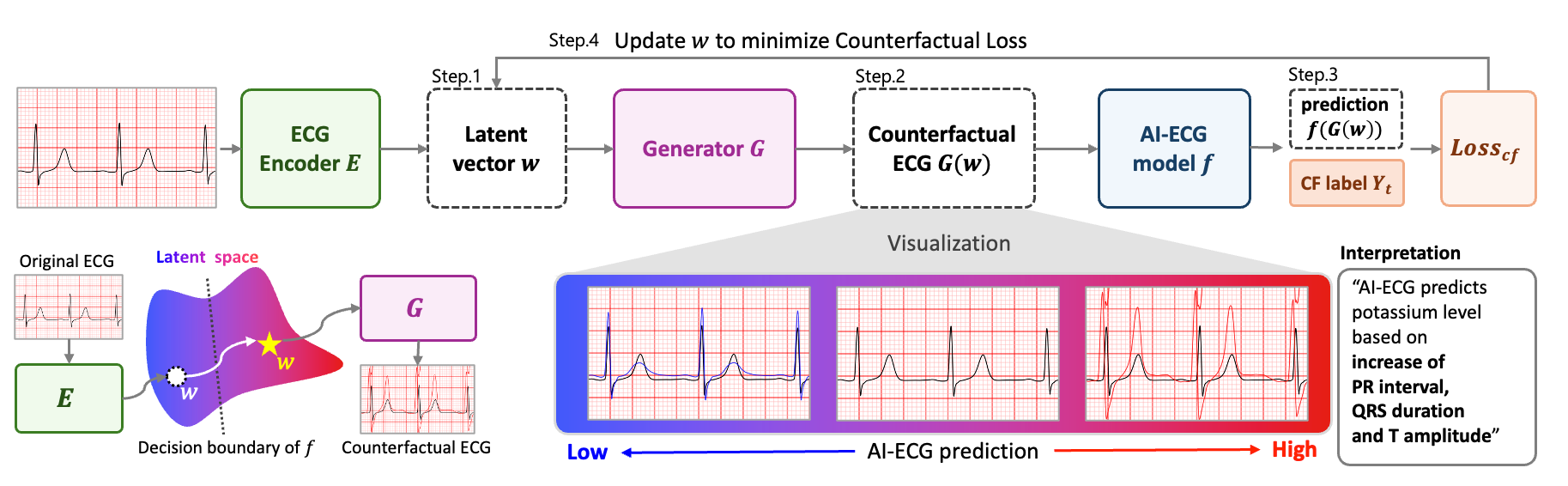} 
    % \vspace{-1mm}
    \caption{
    Process of generating counterfactual ECGs. This example illustrates the key changes in PR interval, QRS duration, and T-wave amplitude by comparing the original and counterfactual ECGs.
    } 
    \label{fig:cf_process}
\end{figure*}

% contribution.
This paper presents three key contributions: (1) we introduce the CoFE, a framework that generate counterfactual ECG including saliency map; (2) we demonstrate the capacity of CoFE to produce clinically coherent explanations; and (3) we show the utility of the CoFE through case studies on Atrial Fibrillation classification and Potassium Level regression models.

\section{CoFE: A Framework for Counterfactual ECG Generation and Interactive Explanation}

The CoFE is designed to enhance the interpretability of AI-based ECG models by generating counterfactual ECGs that explain how specific changes in signal features influence model predictions. CoFE integrates a generative model with a predictive classifier and interactive visualization tools to enable both “where” and “how” explanations for model decisions.

\subsection{Components}\label{sec:components}

At the core of CoFE are three pretrained components: a generator \(G\), an encoder \(E\), and a predictive model \(f\). Together, they form a system that can synthesize clinically plausible ECGs and modify them to explore how small changes in morphology affect diagnostic predictions.

\textbf{Generator }  
\(G\) maps a latent vector \(w \in \mathbb{R}^d\) to a 12-lead ECG signal \(G(w)\). It is implemented using the StyleGAN2 architecture~\cite{karras2020analyzing} and trained with the standard GAN objective~\cite{goodfellow2014generative}:
\[
\min_G \max_D \; \mathbb{E}_{x \sim p_{\text{data}}}[\log D(x)] + \mathbb{E}_{z \sim p_z}[\log (1 - D(G(z)))]
\]
\(D\) is the discriminator trained to distinguish between real and generated ECG signals. Meanwhile, \(G\) is trained to generate ECG signals that \(D\) classifies as real. The generator is trained on 300,000 12-lead ECGs from the MIMIC-IV database~\cite{gow2023mimic}, resampled to 250 Hz, enabling it to synthesize high-fidelity ECG signals that follow real-world cardiac morphology.

\textbf{Encoder }  
\(E\) projects a measured ECG \(x\) into the latent space of \(G\) by learning an inverse mapping \(w = E(x)\). It is trained with a reconstruction objective:
\[
\mathcal{L}_{\text{recon}} = \| x - G(E(x)) \|_2^2
\]
where \(G\) is fixed during training. This ensures that \(E\) can generate latent vectors that accurately reconstruct clinical ECGs through the generator.

\textbf{Predictive model }  
\(f\) is any pretrained AI-ECG model that maps an ECG signal \(x\) to a prediction \(f(x) \in [0,1]^C\), representing probabilities over \(C\) clinical classes. It serves as the target for explanation.

\subsection{Counterfactual ECG Generation}  
Given an input ECG \(x\), we obtain its latent representation \(w_0 = E(x)\), reconstruct the signal as \(\tilde{x}_0 = G(w_0)\), and iteratively update \(w\) to encourage a prediction shift from the original class \(Y = \arg\max f(x)\) to a user-defined target class \(Y_t\). This is achieved by minimizing the counterfactual loss:
\[
\mathcal{L}_{\text{cf}}(w) = \text{CE}(f(G(w)), Y_t)
\]
We update the latent code via gradient descent:
\[
w_{t+1} = w_t - \eta \cdot \nabla_{w_t} \mathcal{L}_{\text{cf}}(w_t)
\]
After \(N\) iterations, we obtain the counterfactual ECG \(\tilde{x}_N = G(w_N)\), which preserves the morphology of the original input while shifting the model prediction toward the target class. Since the optimization occurs in the latent space of a pretrained generator, the resulting ECG remains a plausible sample from the clinical data distribution.

\subsection{Workflow}

\begin{figure*}[!h]
    \centering
    \includegraphics[width=.95\textwidth]{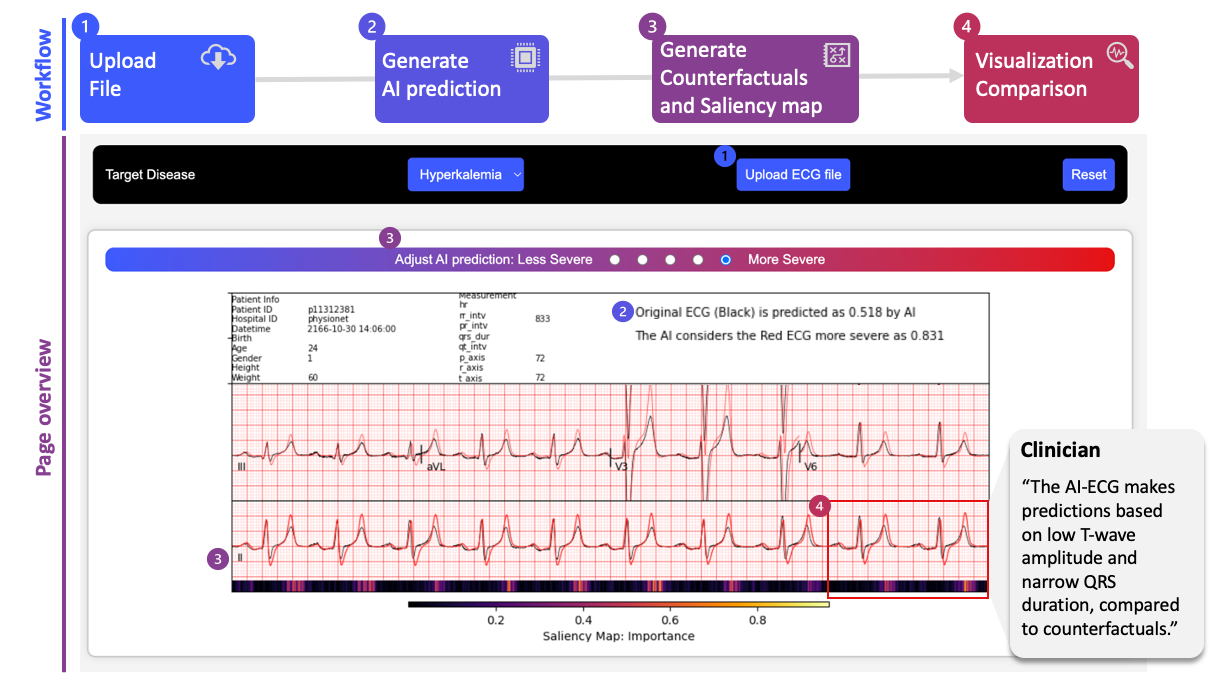} 
    % \vspace{-1mm}
    \caption{
    Overview of the CoFE framework workflow. The diagram illustrates the four key steps: (1) Upload ECG file, (2) Generate AI diagnosis, (3) Generate counterfactual ECG and Saliency Map, and (4) Compare original with counterfactual ECGs.
    } 
    \label{fig:overview}
\end{figure*}

Figure~\ref{fig:overview} presents the four-stage workflow of CoFE:

\begin{enumerate}[leftmargin=*]
  \item \textbf{Upload ECG File}: Clinicians upload a standard-format ECG file (e.g., from Philips or MUSE systems).
  \item \textbf{AI Diagnosis}: The pretrained model \(f\) analyzes the input and produces a classification or regression output.
  \item \textbf{Generate Explanations}: The framework synthesizes counterfactual ECGs that change the model’s output and visualizes saliency maps for the original signal.
  \item \textbf{Compare and Interpret}: The clinician compares the original and counterfactual ECGs side-by-side, with highlighted feature changes and prediction differences.
\end{enumerate}

This workflow enables clinicians to observe how changes in specific intervals (e.g., RR variability) or amplitudes (e.g., P- or T-wave amplitude) affect predictions.

\subsection{Key Features}

\textbf{Integrated Explanation.} CoFE combines counterfactual ECGs with saliency maps, providing insight into both the spatial (saliency) and causal (counterfactual) aspects of the model’s reasoning.

\noindent\textbf{Interactive Controls.} Clinicians can interactively explore counterfactual scenarios using intuitive sliders to control the severity of modifications, making the system suitable for hypothesis testing or educational purposes.

\noindent\textbf{Clinically Familiar UI.} Counterfactual ECGs are overlaid on original signals using conventional ECG layouts. Changes are marked in color (e.g., red overlays), which reduces cognitive load and enhances clinician trust.

% \subsection{Illustrative Use Case}

Figure~\ref{fig:overview} illustrates a case where CoFE is applied to an atrial fibrillation classifier. The original ECG displays regular P-waves and low RR variability. The counterfactual version exhibits suppressed P-wave amplitude and increased RR variability, thereby shifting the model's prediction toward atrial fibrillation (AF). Saliency maps confirm that the model focuses on these modified regions.

This example illustrates CoFE’s ability to link model behavior with clinically meaningful changes in signal features, providing actionable insights into both model interpretation and potential clinical scenarios.

\begin{figure*}[!h]
    \centering
    \includegraphics[width=0.95\textwidth]{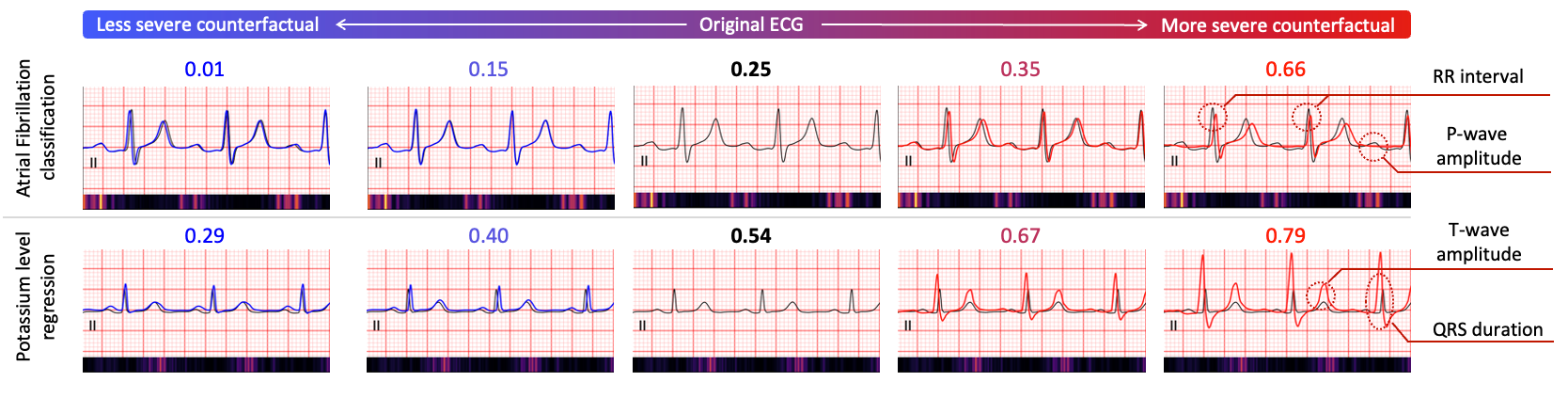} 
    % \vspace{-1mm}
    \caption{
    Comparison of original and counterfactual ECGs generated by the CoFE framework: In the top row (AF classification), the values above each ECG show the model’s predicted probability of AF. As the number increases, the model becomes more confident that the ECG indicates atrial fibrillation (AF). In the bottom row (potassium level regression), the values represent the model’s prediction of the potassium level, normalized to a range of 0 to 1. Higher values indicate a higher predicted potassium level.
    } 
    \label{fig:casestudy}
\end{figure*}

\section{Demo Scenario}
% For demo, we choose two AI-diagnostic models: (1) an atrial fibrillation (AF) classification model~\cite{clifford2017af} and (2) a potassium level regression model~\cite{jang2024unveiling}. Each model is trained from scratch with the Physionet challenge 2017 dataset~\cite{clifford2017af}.

For demonstration, we apply CoFE to interpret two AI-based diagnostic models, which correspond to the predictive model \(f\) introduced in Section~\ref{sec:components}. Both models are trained on 12-lead ECG recordings using 10-second segments sampled at 250~Hz, and share the same backbone architecture—a 1D adaptation of ResNet-16, where 2D convolutions are replaced with 1D operations for ECG modeling. Training is performed using the Adam optimizer (learning rate: 0.001) with early stopping based on validation loss.

The models differ in their task objectives, label sources, and loss functions:

\begin{itemize}
    \item \textbf{AF classification model}: This model is trained from scratch on the PhysioNet/Computing in Cardiology Challenge 2017 dataset~\cite{clifford2017af} to perform binary classification of atrial fibrillation. It is optimized using cross-entropy loss with the diagnostic labels provided in the dataset.

    \item \textbf{Potassium regression model}: This model estimates continuous serum potassium concentrations and follows the training procedure in~\cite{jang2024unveiling}. It is trained on ECG–potassium pairs obtained from the MIMIC-IV database~\cite{johnson2016mimic}, using mean squared error (MSE) loss.
\end{itemize}

To evaluate CoFE, we randomly sample 1,000 ECG recordings from the MIMIC-IV dataset~\cite{johnson2016mimic}. These evaluation samples are strictly disjoint from those used to train the potassium regression model, ensuring an unbiased assessment. We assess the statistical significance of counterfactual modifications and present representative examples to illustrate practical interpretability. 

A demonstration video illustrating these scenarios is available at: \url{https://www.youtube.com/watch?v=YoW0bNBPglQ}.

\subsection{Clinical Background}

Atrial fibrillation (AF) is a common cardiac arrhythmia marked by rapid and disorganized electrical activity in the atria. Its hallmark ECG features include \textbf{irregular RR intervals} and \textbf{diminished or absent P-waves}~\cite{bollmann2006analysis}. Detecting these subtle changes is essential for accurate diagnosis and timely intervention.

Potassium is a key electrolyte that directly affects cardiac electrophysiology. Elevated serum potassium levels typically present as \textbf{increased T-wave amplitude} and \textbf{prolonged QRS duration} on the ECG~\cite{akbilgic2021ecg}. Early recognition of these patterns is critical for the diagnosis and management of electrolyte imbalances such as hyperkalemia.

\section{Evaluation}
\subsection{Quantitative Result}
To quantitatively evaluate the counterfactual ECGs, we applied CoFE to generate positive counterfactuals, i.e., synthetic ECGs that the model interprets as reflecting more severe clinical states than the original inputs. These counterfactuals were then compared with the original ECGs to assess whether key physiological features were altered in clinically meaningful ways (Table~\ref{tab:feature_analysis}).

\begin{table}[htbp]
\centering
\footnotesize
\caption{Statistical comparison of ECG Features from original ECGs and severed counterfactual ECGs.}
\setlength{\tabcolsep}{1pt}
\renewcommand{\arraystretch}{2.0}
\begin{tabular}{|c|c|c|c|c|}
\hline
\rowcolor[HTML]{EFEFEF} 
\textbf{AI-ECG} & \textbf{Feature} &\textbf{ \makecell{Original \\ ECGs}} & \textbf{\makecell{Counterfactual \\ ECGs}} & \textbf{p-value} \\ \hline
\multirow{2}{*}{Atrial Fibrillation} & P-wave amplitude (mV) & 0.09 & 0.04 & $<$0.05 \\ \cline{2-5} 
 & RR interval variability (ms) & 46.17 & 85.44 & $<$0.05 \\ \hline
\multirow{2}{*}{Potassium level} & T-wave amplitude (mV) & 0.18 & 0.58 & $<$0.05 \\ \cline{2-5} 
 & QRS duration (ms) & 95.84 & 121.98 & $<$0.05 \\ \hline
\end{tabular}
\label{tab:feature_analysis}

\end{table}

For the atrial fibrillation (AF) classification task, CoFE generated counterfactuals that showed a marked \textbf{reduction in P-wave amplitude} (from 0.09 mV to 0.04 mV) and an \textbf{increase in RR interval variability} (from 46.17 ms to 85.44 ms). These changes align with established clinical signatures of AF, reflecting diminished atrial activity and irregular heart rhythms.

For the potassium level regression task, CoFE-generated counterfactuals exhibited \textbf{increased T-wave amplitude} (from 0.18 mV to 0.58 mV) and \textbf{prolonged QRS duration} (from 95.84 ms to 121.98 ms). These findings are consistent with known ECG manifestations of hyperkalemia, including enhanced repolarization and slowed ventricular conduction.

These results demonstrate that the proposed framework generates physiologically plausible counterfactuals that align with established clinical knowledge, supporting its utility for model interpretability and validation.

\subsection{Qualitative Result}

Figure~\ref{fig:casestudy} illustrates how CoFE-generated counterfactuals complement saliency-based interpretations for two distinct tasks.

In the AF classification task (top row), the saliency map highlights the P-wave region, suggesting that the model relies heavily on atrial activity. The counterfactual ECGs show a reduced P-wave amplitude and increased RR interval variability, both of which consistently push the model's prediction toward AF. This indicates that the model has learned clinically relevant patterns, including diminished atrial depolarization and irregular rhythm.

In the potassium level regression task (bottom row), the saliency map focuses on the T-wave, indicating its importance in estimating serum potassium. The corresponding counterfactuals exhibit increased T-wave amplitude and prolonged QRS duration, both of which elevate the predicted potassium level. These changes align with well-established ECG manifestations of hyperkalemia.

Together, the saliency maps and counterfactuals provide complementary insights, confirming that the model's predictions are grounded in physiologically meaningful features.

\section{Conclusion}
We introduced the CoFE framework, designed to enhance the interpretability of AI-based ECG models by coupling counterfactual ECGs with saliency maps. By illustrating both where a model focuses and why certain features matter, our framework provides a more comprehensive explanation than attribution methods alone. Our case studies on AF classification and potassium level regression demonstrate that CoFE framework identifies clinically meaningful changes. This integrated view helps clinicians understand the relationships underlying AI-ECG's decisions, improving confidence and guiding more informed clinical judgments. Beyond these initial applications, the CoFE framework can be extended to broader biosignal domains. In future work, integrating large language models could offer narrative-style explanations, further enhancing accessibility and trustworthiness. 

\newpage
% \section*{Acknowledgments}
% Thank you for your submission to the European Heart Journal. We are delighted to inform you that your paper entitled "Prospective Multi-center External Validation of Rule-Out acute Myocardial Infarction using Artificial intelligence Electrocardiogram analysis (ROMIAE)" with reference number EURHEARTJ-D-24-02682R2 has been accepted for publication.

\bibliographystyle{ACM-Reference-Format}
\bibliography{sample-base}


\begin{thebibliography}{13}

%%% the Web address.

\ifx \showCODEN    \undefined \def \showCODEN     #1{\unskip}     \fi
\ifx \showDOI      \undefined \def \showDOI       #1{#1}\fi
\ifx \showISBNx    \undefined \def \showISBNx     #1{\unskip}     \fi
\ifx \showISBNxiii \undefined \def \showISBNxiii  #1{\unskip}     \fi
\ifx \showISSN     \undefined \def \showISSN      #1{\unskip}     \fi
\ifx \showLCCN     \undefined \def \showLCCN      #1{\unskip}     \fi
\ifx \shownote     \undefined \def \shownote      #1{#1}          \fi
\ifx \showarticletitle \undefined \def \showarticletitle #1{#1}   \fi
\ifx \showURL      \undefined \def \showURL       {\relax}        \fi
% invisible to TeX
\providecommand\bibfield[2]{#2}
\providecommand\bibinfo[2]{#2}
\providecommand\natexlab[1]{#1}
\providecommand\showeprint[2][]{arXiv:#2}

\bibitem[Akbilgic et~al\mbox{.}(2021)]%
        {akbilgic2021ecg}
\bibfield{author}{\bibinfo{person}{Oguz Akbilgic}, \bibinfo{person}{Liam Butler}, \bibinfo{person}{Ibrahim Karabayir}, \bibinfo{person}{Patricia~P Chang}, \bibinfo{person}{Dalane~W Kitzman}, \bibinfo{person}{Alvaro Alonso}, \bibinfo{person}{Lin~Y Chen}, {and} \bibinfo{person}{Elsayed~Z Soliman}.} \bibinfo{year}{2021}\natexlab{}.
\newblock \showarticletitle{ECG-AI: electrocardiographic artificial intelligence model for prediction of heart failure}.
\newblock \bibinfo{journal}{\emph{European Heart Journal-Digital Health}} (\bibinfo{year}{2021}).
\newblock


\bibitem[Attia et~al\mbox{.}(2021)]%
        {attia2021application}
\bibfield{author}{\bibinfo{person}{Zachi~I Attia}, \bibinfo{person}{David~M Harmon}, \bibinfo{person}{Elijah~R Behr}, {and} \bibinfo{person}{Paul~A Friedman}.} \bibinfo{year}{2021}\natexlab{}.
\newblock \showarticletitle{Application of artificial intelligence to the electrocardiogram}.
\newblock \bibinfo{journal}{\emph{European heart journal}} (\bibinfo{year}{2021}).
\newblock


\bibitem[Bollmann et~al\mbox{.}(2006)]%
        {bollmann2006analysis}
\bibfield{author}{\bibinfo{person}{Andreas Bollmann}, \bibinfo{person}{Daniela Husser}, \bibinfo{person}{Luca Mainardi}, \bibinfo{person}{Federico Lombardi}, \bibinfo{person}{Philip Langley}, \bibinfo{person}{Alan Murray}, \bibinfo{person}{Jose~Joaquin Rieta}, \bibinfo{person}{Jose Millet}, \bibinfo{person}{S~Bertil Olsson}, \bibinfo{person}{Martin Stridh}, {et~al\mbox{.}}} \bibinfo{year}{2006}\natexlab{}.
\newblock \showarticletitle{Analysis of surface electrocardiograms in atrial fibrillation: techniques, research, and clinical applications}.
\newblock \bibinfo{journal}{\emph{Europace}} (\bibinfo{year}{2006}).
\newblock


\bibitem[Clifford et~al\mbox{.}(2017)]%
        {clifford2017af}
\bibfield{author}{\bibinfo{person}{Gari~D Clifford}, \bibinfo{person}{Chengyu Liu}, \bibinfo{person}{Benjamin Moody}, \bibinfo{person}{Li-wei~H Lehman}, \bibinfo{person}{Ikaro Silva}, \bibinfo{person}{Qiao Li}, \bibinfo{person}{AE Johnson}, {and} \bibinfo{person}{Roger~G Mark}.} \bibinfo{year}{2017}\natexlab{}.
\newblock \showarticletitle{AF classification from a short single lead ECG recording: The PhysioNet/computing in cardiology challenge 2017}. In \bibinfo{booktitle}{\emph{2017 computing in cardiology (CinC)}}.
\newblock


\bibitem[Goodfellow et~al\mbox{.}(2014)]%
        {goodfellow2014generative}
\bibfield{author}{\bibinfo{person}{Ian Goodfellow}, \bibinfo{person}{Jean Pouget-Abadie}, \bibinfo{person}{Mehdi Mirza}, \bibinfo{person}{Bing Xu}, \bibinfo{person}{David Warde-Farley}, \bibinfo{person}{Sherjil Ozair}, \bibinfo{person}{Aaron Courville}, {and} \bibinfo{person}{Yoshua Bengio}.} \bibinfo{year}{2014}\natexlab{}.
\newblock \showarticletitle{Generative adversarial nets}.
\newblock \bibinfo{journal}{\emph{Advances in neural information processing systems}} (\bibinfo{year}{2014}).
\newblock


\bibitem[Gow et~al\mbox{.}(2023)]%
        {gow2023mimic}
\bibfield{author}{\bibinfo{person}{Brian Gow}, \bibinfo{person}{Tom Pollard}, \bibinfo{person}{Larry~A Nathanson}, \bibinfo{person}{Alistair Johnson}, \bibinfo{person}{Benjamin Moody}, \bibinfo{person}{Chrystinne Fernandes}, \bibinfo{person}{Nathaniel Greenbaum}, \bibinfo{person}{Seth Berkowitz}, \bibinfo{person}{Dana Moukheiber}, \bibinfo{person}{Parastou Eslami}, {et~al\mbox{.}}} \bibinfo{year}{2023}\natexlab{}.
\newblock \showarticletitle{Mimic-iv-ecg-diagnostic electrocardiogram matched subset}.
\newblock \bibinfo{journal}{\emph{Type: dataset}} (\bibinfo{year}{2023}).
\newblock


\bibitem[Jang et~al\mbox{.}(2024)]%
        {jang2024unveiling}
\bibfield{author}{\bibinfo{person}{Jong-Hwan Jang}, \bibinfo{person}{Yong-Yeon Jo}, \bibinfo{person}{Sora Kang}, \bibinfo{person}{Jeong~Min Son}, \bibinfo{person}{Hak~Seung Lee}, \bibinfo{person}{Joon-myoung Kwon}, {and} \bibinfo{person}{Min~Sung Lee}.} \bibinfo{year}{2024}\natexlab{}.
\newblock \showarticletitle{Unveiling AI-ECG using Generative Counterfactual XAI Framework}.
\newblock \bibinfo{journal}{\emph{medRxiv}} (\bibinfo{year}{2024}), \bibinfo{pages}{2024--09}.
\newblock


\bibitem[Jo et~al\mbox{.}(2021)]%
        {jo2021detection}
\bibfield{author}{\bibinfo{person}{Yong-Yeon Jo}, \bibinfo{person}{Joon-myoung Kwon}, \bibinfo{person}{Ki-Hyun Jeon}, \bibinfo{person}{Yong-Hyeon Cho}, \bibinfo{person}{Jae-Hyun Shin}, \bibinfo{person}{Yoon-Ji Lee}, \bibinfo{person}{Min-Seung Jung}, \bibinfo{person}{Jang-Hyeon Ban}, \bibinfo{person}{Kyung-Hee Kim}, \bibinfo{person}{Soo~Youn Lee}, {et~al\mbox{.}}} \bibinfo{year}{2021}\natexlab{}.
\newblock \showarticletitle{Detection and classification of arrhythmia using an explainable deep learning model}.
\newblock \bibinfo{journal}{\emph{Journal of Electrocardiology}} (\bibinfo{year}{2021}).
\newblock


\bibitem[Johnson et~al\mbox{.}(2016)]%
        {johnson2016mimic}
\bibfield{author}{\bibinfo{person}{Alistair~EW Johnson}, \bibinfo{person}{Tom~J Pollard}, \bibinfo{person}{Lu Shen}, \bibinfo{person}{Li-wei~H Lehman}, \bibinfo{person}{Mengling Feng}, \bibinfo{person}{Mohammad Ghassemi}, \bibinfo{person}{Benjamin Moody}, \bibinfo{person}{Peter Szolovits}, \bibinfo{person}{Leo Anthony~Celi}, {and} \bibinfo{person}{Roger~G Mark}.} \bibinfo{year}{2016}\natexlab{}.
\newblock \showarticletitle{MIMIC-III, A Freely Accessible Critical Care Database}.
\newblock \bibinfo{journal}{\emph{Scientific Data}} (\bibinfo{year}{2016}).
\newblock


\bibitem[Karras et~al\mbox{.}(2020)]%
        {karras2020analyzing}
\bibfield{author}{\bibinfo{person}{Tero Karras}, \bibinfo{person}{Samuli Laine}, \bibinfo{person}{Miika Aittala}, \bibinfo{person}{Janne Hellsten}, \bibinfo{person}{Jaakko Lehtinen}, {and} \bibinfo{person}{Timo Aila}.} \bibinfo{year}{2020}\natexlab{}.
\newblock \showarticletitle{Analyzing and improving the image quality of stylegan}. In \bibinfo{booktitle}{\emph{Proceedings of the IEEE/CVF conference on computer vision and pattern recognition}}.
\newblock


\bibitem[Niebur(2007)]%
        {niebur2007saliency}
\bibfield{author}{\bibinfo{person}{Ernst Niebur}.} \bibinfo{year}{2007}\natexlab{}.
\newblock \showarticletitle{Saliency map}.
\newblock \bibinfo{journal}{\emph{Scholarpedia}} (\bibinfo{year}{2007}).
\newblock


\bibitem[Selvaraju et~al\mbox{.}(2017)]%
        {selvaraju2017grad}
\bibfield{author}{\bibinfo{person}{Ramprasaath~R Selvaraju}, \bibinfo{person}{Michael Cogswell}, \bibinfo{person}{Abhishek Das}, \bibinfo{person}{Ramakrishna Vedantam}, \bibinfo{person}{Devi Parikh}, {and} \bibinfo{person}{Dhruv Batra}.} \bibinfo{year}{2017}\natexlab{}.
\newblock \showarticletitle{Grad-cam: Visual explanations from deep networks via gradient-based localization}. In \bibinfo{booktitle}{\emph{Proceedings of the IEEE international conference on computer vision}}.
\newblock


\bibitem[Verma et~al\mbox{.}(2020)]%
        {verma2020counterfactual}
\bibfield{author}{\bibinfo{person}{Sahil Verma}, \bibinfo{person}{John Dickerson}, {and} \bibinfo{person}{Keegan Hines}.} \bibinfo{year}{2020}\natexlab{}.
\newblock \showarticletitle{Counterfactual explanations for machine learning: A review}.
\newblock \bibinfo{journal}{\emph{arXiv preprint arXiv:2010.10596}} (\bibinfo{year}{2020}).
\newblock


\end{thebibliography}

%%
% %% If your work has an appendix, this is the place to put it.
% \appendix





\end{document}